# Assessing AI-Generated Questions' Alignment with Cognitive Frameworks in Educational Assessment

Antoun Yaacoub[1,], Jérôme Da-Rugna[1] and Zainab Assaghir[2]

[1]Learning, Data and Robotics (LDR) ESIEA Lab, ESIEA, Paris, France
[2]Faculty of Science, Lebanese University, Beirut, Lebanon
Email: antoun.yaacoub@esiea.fr (A.Y.); jerome.darugna@esiea.fr (J.D.R.); zainab.assaghir@ul.edu.lb (Z.A.)

*Abstract*—This study evaluates the integration of Bloom's Taxonomy into OneClickQuiz, an Artificial Intelligence (AI) driven plugin for automating Multiple-Choice Question (MCQ) generation in Moodle. Bloom's Taxonomy provides a structured framework for categorizing educational objectives into hierarchical cognitive levels. Our research investigates whether incorporating this taxonomy can improve the alignment of AI-generated questions with specific cognitive objectives. We developed a dataset of 3691 questions categorized according to Bloom's levels and employed various classification models—Multinomial Logistic Regression, Naive Bayes, Linear Support Vector Classification (SVC), and a Transformer-based model (DistilBERT)—to evaluate their effectiveness in categorizing questions. Our results indicate that higher Bloom's levels generally correlate with increased question length, Flesch-Kincaid Grade Level (FKGL), and Lexical Density (LD), reflecting the increased complexity of higher cognitive demands. Multinomial Logistic Regression showed varying accuracy across Bloom's levels, performing best for "Knowledge" and less accurately for higher-order levels. Merging higher-level categories improved accuracy for complex cognitive tasks. Naive Bayes and Linear SVC also demonstrated effective classification for lower levels but struggled with higher-order tasks. DistilBERT achieved the highest performance, significantly improving classification of both lower and higher-order cognitive levels, achieving an overall validation accuracy of 91%. This study highlights the potential of integrating Bloom's Taxonomy into AI-driven assessment tools and underscores the advantages of advanced models like DistilBERT for enhancing educational content generation.

*Keywords*—artificial intelligence, machine learning, educational technology, bloom taxonomy, education system and its application, natural language processing.

## I. INTRODUCTION

Educational assessment frameworks are foundational in shaping how educators evaluate student learning processes and outcomes. Bloom's Taxonomy [1], introduced in 1956 by Benjamin Bloom and his collaborators, is one of the most influential frameworks in educational theory. Its hierarchical structure organizes cognitive learning objectives into six levels: Knowledge, Comprehension, Application, Analysis, Synthesis, and Evaluation. Each level represents increasingly complex cognitive processes, guiding educators in designing curricula, instructional methods, and assessment strategies. The taxonomy's continued relevance reflects its versatility across disciplines and educational settings [2].

However, contemporary education, driven by technological advances and evolving pedagogical paradigms, requires new methods for integrating such frameworks into teaching and assessment. Artificial intelligence (AI), particularly generative AI, offers promising solutions by automating complex tasks like question generation and content creation [3]. AI has gained considerable traction in educational technology, enabling tools that support personalized learning, adaptive assessments, and efficient resource generation. Generative AI models such as GPT-4 have been used to create quizzes, essays, and other instructional materials, showcasing their potential to transform traditional education [4, 5].

Despite these advancements, the alignment of AI-generated content with established educational frameworks like Bloom's Taxonomy remains under-explored. While AI-driven assessment tools generate varied and contextually relevant questions, their adherence to hierarchical cognitive levels hasn't been rigorously assessed. Integrating Bloom's Taxonomy into AI-based question generation ensures that AI-generated assessments are pedagogically sound and aligned with educators' cognitive objectives [6].

This study explores whether integrating Bloom's Taxonomy into OneClickQuiz, an AI-powered Moodle plugin for automated MCQ generation, enhances the alignment of AI-generated questions with specific cognitive levels. OneClickQuiz has been used in prior research [7] to generate diverse quiz content for Moodle, but it lacked a systematic approach to aligning questions with Bloom's cognitive framework. This study investigates the extent to which generative AI can produce questions reflecting the taxonomy's structured cognitive demands.

We address two primary research questions:

RQ1: To what extent can AI-generated questions, categorized according to Bloom's Taxonomy, accurately represent different cognitive levels?

RQ2: How effective are different AI models, including traditional classification algorithms and advanced Transformer-based models, in classifying and generating questions aligned with the higher-order cognitive processes outlined in Bloom's Taxonomy?

The findings will inform the design of AI-driven educational tools, offering insights into refining AI models to support assessments promoting critical thinking and higher-order learning. By integrating Bloom's Taxonomy into OneClickQuiz, we contribute to discussions about AI's role in enhancing the quality and cognitive depth of automated assessments. This work provides both theoretical insights and practical recommendations for improving AI-driven assessment technologies.

The paper is organized as follows: Section II presents the literature review; Section III details the methodology and experimental setup; Section IV presents and analyzes the results; and Section V discusses the implications and concludes with recommendations for future research.

## II. LITERATURE REVIEW

This section provides a comprehensive overview of Bloom's Taxonomy and other relevant educational taxonomies, including Webb's Depth of Knowledge (DOK), the Structure of Observed Learning Outcomes (SOLO) Taxonomy, and Bloom's Digital Taxonomy. It also explores the integration of AI and generative AI in education, highlighting their impact on quiz and MCQ creation, and discusses ethical considerations and challenges associated with AI in educational contexts.

### A. Bloom's Taxonomy and Other Educational Taxonomies

Bloom's Taxonomy [1], first published in 1956 by Benjamin Bloom and colleagues, remains a seminal framework in educational theory and practice. It organizes cognitive learning objectives into a hierarchical structure with six levels, each representing different degrees of complexity:

1. **Knowledge**: This foundational level involves the recall of factual information, basic concepts, and definitions. Objectives at this level include listing, naming, and recalling information.
2. **Comprehension**: At this level, learners are expected to understand and interpret information. This includes explaining, summarizing, and discussing concepts.
3. **Application**: This level requires learners to apply knowledge to new situations or problems. Objectives include using information in practical contexts, solving problems, and making decisions.
4. **Analysis**: Learners at this level are expected to break down information into its constituent parts and understand relationships. This involves differentiating, analyzing, and examining components.
5. **Synthesis**: This higher-order level focuses on combining parts to form a new whole. Objectives include creating, designing, and proposing new solutions or structures.
6. **Evaluation**: The highest level involves making judgments about the value or quality of information or solutions. Objectives include critiquing, evaluating, and justifying decisions.

Bloom's Taxonomy has been widely used in curriculum development, instructional design, and assessment. It helps educators create learning objectives that promote higher-order thinking and ensures that assessments align with learning goals.

In 2001, Lorin Anderson and David Krathwohl [2, 8] revised Bloom's Taxonomy to reflect contemporary educational practices. The revised taxonomy introduced several changes:

- **Revised Categories**: The original categories were updated to reflect active learning processes. The revised levels are Remembering, Understanding, Applying, Analyzing, Evaluating, and Creating.
- **Cognitive Process Dimension**: The revised taxonomy emphasizes cognitive processes, such as recalling (Remembering), explaining (Understanding), and designing (Creating), reflecting a more dynamic view of learning.
- **Knowledge Dimension**: The revised taxonomy also includes a Knowledge Dimension, which classifies knowledge into Factual, Conceptual, Procedural, and Metacognitive categories.

Webb's Depth of Knowledge (DOK) [9], developed by Norman Webb, provides an alternative framework for categorizing cognitive tasks. It focuses on the complexity of tasks rather than hierarchical levels:

1. **Level 1**: Recall and Reproduction—Simple recall of facts and basic skills.
2. **Level 2**: Skills and Concepts—Applying skills and concepts to solve problems.
3. **Level 3**: Strategic Thinking—Higher-order thinking requiring reasoning and planning.
4. **Level 4**: Extended Thinking—Complex tasks that involve extended research and investigation.

Webb's DOK framework emphasizes the depth of understanding required for various cognitive tasks, providing a different perspective on instructional design and assessment.

SOLO's Taxonomy [10], developed by John Biggs and Kevin Collis, categorizes learning outcomes into levels of complexity, including Pre-Structural, Uni-Structural, Multi-Structural, Relational, and Extended Abstract.

Bloom's Digital Taxonomy is an adaptation of Bloom's Taxonomy that incorporates digital tools and technologies. It reflects the impact of digital media on learning processes and includes categories such as Remembering with digital tools, Understanding through multimedia, and Creating with digital content.

We chose Bloom's Taxonomy over newer frameworks due to its established and widely recognized hierarchical structure, which provides a clear, time-tested method for categorizing cognitive objectives and assessing learning outcomes. Its broad acceptance and historical significance in educational practice make it an ideal benchmark for evaluating the effectiveness of AI-generated questions.

### B. The Integration of AI and Generative AI in Education

The following subsection investigates the integration of generative AI in educational contexts, focusing on advancements, limitations, and the influence on assessment and curriculum development, particularly as it relates to our OneClickQuiz plugin.

Generative AI technologies, such as Large Language Models (LLMs) like GPT-4 and PALM2, are revolutionizing educational practices by enhancing quiz and MCQ creation. Eager and Brunton [11] and Trust et al. [3] highlight the transformative impact of these models on teaching, assessment, and curriculum design. Pack and Maloney [12] demonstrate how OpenAI's ChatGPT assists in language education through effective information compilation and summarization, while Zhai and Nehm [13] argue for a broader application of AI in formative assessment, advocating its use for feedback and diversified assessment methods.

Recent research underscores the importance of understanding factors influencing the adoption of AI tools in education. A study on the determinants of AI application usage among Humanities and Social Sciences students reveals that expected performance, habit, and enjoyment are key factors influencing students' intentions to use AI applications [14]. This finding is crucial for ensuring that AI tools like OneClickQuiz meet users' needs and expectations.

Generative AI's role in personalizing education is further exemplified by Olga et al. [15], who explore its application in tailoring learning experiences. Doughty et al. [16] illustrate GPT-4's potential in generating high-quality, contextually relevant programming MCQs, while Van Campenhout et al. [17] assess AI-generated questions in Psychology, considering student data and perceptions. Additionally, a study on how teaching interventions improve students' information-seeking behaviors demonstrates the value of targeted educational support in enhancing AI tool effectiveness [18].

A comprehensive book on innovative technologies in education, including AI, IoT, and ICT, highlights various applications and case studies showcasing how these technologies are integrated into educational settings to address diverse needs [6].

These discussions underscore AI's potential to produce engaging and pertinent educational content, aligning with the objectives of our OneClickQuiz plugin for automated quiz generation.

Despite the advancements, several challenges persist. Akgun and Greenhow [19] emphasize the importance of addressing ethical considerations in AI applications within K-12 education. Dai and Ke [20] stress the need for careful integration of AI with educational principles in simulation-based learning, while Bahroun et al. [21] point out concerns regarding bias, transparency, and equity in AI tools. These ethical issues highlight the necessity for responsible AI deployment, ensuring fairness and transparency in educational tools like our plugin.

Moreover, a recent comprehensive review of AI in primary education illustrates the broad scope of AI applications, categorizing research objectives, learning content, activities, and outcomes [22]. This review emphasizes the diverse pedagogical approaches and learning outcomes facilitated by AI tools, reinforcing the need for AI applications to align with established educational frameworks like Bloom's Taxonomy.

AI's potential to enhance assessment methods and curriculum development is significant. Grévisse [23] assesses the alignment of GPT-based MCQs with best practices, while Jain [24] explores automated quiz generation for programming languages.

Ryan Lau [25] developed a task complexity classifier using Transformer-based NLP model based on Bloom's Taxonomy. Lau deployed an architecture where users can submit a query as a string, and the application returns the corresponding Bloom's category.

*C. Integrating AI and Bloom's Taxonomy in Educational Technology*

In the digital age, the integration of AI with Bloom's Taxonomy has emerged as a significant research focus, given its potential to transform educational practices, especially in assessment and personalized learning.

Recent advancements, such as the BloomGPT project [26], have demonstrated how AI tools like ChatGPT can be integrated with Bloom's Taxonomy to support cognitive development across multiple educational levels. In a university setting, BloomGPT facilitated discussions and reflections that improved students' conceptual, procedural, and metacognitive knowledge, highlighting the utility of AI in enhancing learning through Socratic dialogues and essay evaluations.

The AIEd Bloom's Taxonomy model [27] presents an AI-driven framework designed to enhance the effectiveness of educational tools. This model adapts Bloom's Taxonomy to modern AI capabilities, incorporating levels like collecting, processing, and innovating to align with the digital learning environment. This approach not only augments learning efficiency but also aligns AI functionalities with Bloom's cognitive levels, offering personalized and adaptive learning experiences.

In design education, AI tools have started to challenge traditional learning sequences. Studies on AI in robotic design projects reveal that students sometimes engage in higher-order tasks, like *Creating*, before mastering foundational knowledge. This potential shift in learning paradigms calls for a re-evaluation of how Bloom's Taxonomy might be applied in AI-driven environments, where the traditional progression of cognitive skills may be reversed [28]. Such findings suggest that AI might require adjustments to fit pedagogical models that prioritize foundational skills before advanced cognitive tasks.

In primary education, the integration of Bloom's Taxonomy in AI-supported learning environments has shed light on how AI tools influence student cognition [29]. For instance, students using chatbots tended to rely on higher-order skills, such as *Evaluating* and *Creating*, during tasks involving complex problem-solving. However, in scenarios of increased task uncertainty, they reverted to lower-order thinking, such as *Remembering* and *Understanding*.

AI tools like SACITED [30], which combines AI with Bloom's Digital Taxonomy, have shown promise in creating didactic sequences for various educational disciplines. SACITED enhances learning by personalizing content and adapting it to the cognitive needs of learners, demonstrating the broad potential of AI to align with and extend the capabilities of Bloom's framework across diverse fields.

Further exploration of AI in educational assessments includes research on automated Multiple-Choice Question (MCQ) generation using GPT-4 aligned with Bloom's Taxonomy [31]. This study focuses on MCQ generation through a few-shot prompting approach and evaluates how well the generated questions align with Bloom's cognitive framework. A key challenge identified is GPT-4's difficulty in generating questions that target higher-order cognitive skills such as Analyzing and Evaluating. The research highlights the limitations of GPT-4 in aligning with these advanced cognitive demands.

Recent research has also applied Bloom's Taxonomy to job tasks in AI-related fields. A semi-supervised model for AI-related job tasks demonstrated the taxonomy's relevance beyond educational contexts, where it was used to classify technical and cognitive requirements [32]. This approach showed how Bloom's hierarchical model could assist in analyzing the complexity of tasks in emerging fields like AI, further reinforcing the taxonomy's adaptability.

These studies underscore the necessity of integrating Bloom's Taxonomy into AI-driven educational tools, both to ensure pedagogical soundness and to enhance learning efficiency. The combination of AI with structured cognitive frameworks offers a new dimension for personalized, adaptive learning, although further refinement is required for AI to fully capture the complexities of higher-order thinking as outlined by Bloom, supporting the development of a model that bridges existing gaps in AI-generated assessments.

### III. MATERIALS AND METHODS

This section details the methodology used for generating and evaluating AI-generated questions. It describes the data creation process using Google's Vertex AI, the metrics used for evaluating question quality, and the classification techniques employed, including Multinomial Logistic Regression, Naive Bayes, Linear Support Vector Classification (SVC), and Transformer-Based Deep Learning Models.

#### A. Data Creation and Description

To construct a comprehensive dataset for evaluating AI-generated questions in relation to Bloom's Taxonomy, we utilized Google's Vertex AI's Text Generation Model (text-bison-32k), chosen for its ability to generate contextually relevant and diverse questions based on pre-defined prompts. The sampling of generated questions focused on the six cognitive levels of Bloom's Taxonomy within the field of computer science, ensuring coverage of each cognitive level: Knowledge, Comprehension, Application, Analysis, Synthesis, and Evaluation. The action verbs were selected from established references [2] and resources from educational institutions such as Iowa State University's Revised Bloom's Taxonomy [1] and Andrew Churches' Bloom's Digital Taxonomy [33].

Survey Design: The design of the questions for each cognitive level was carefully curated. For each level of Bloom's Taxonomy, we incorporated specific action verbs commonly associated with that level to guide the question generation. The following examples outline how questions were formulated for each construct:

- Knowledge: Questions aimed at recalling basic facts or definitions (e.g., "Define the primary functions of an operating system"). Verbs like "define," "list," and "describe" were used to ensure questions focused on simple recall.
- Comprehension: Questions requiring students to explain concepts in their own words (e.g., "Summarize the differences between Transmission Control Protocol (TCP) and User Datagram Protocol (UDP)"). Verbs such as "summarize," "explain," and "interpret" were chosen.
- Application: For this level, the questions focused on applying knowledge to new situations (e.g., "Apply the binary search algorithm to the following list of numbers"). Verbs like "apply" and "solve" were utilized.
- Analysis: Questions required breaking down concepts to understand their components (e.g., "Analyze the components of a cloud computing system"). Verbs like "analyze" and "differentiate" guided the prompts.
- Synthesis: Questions at this level asked for the creation of something new (e.g., "Propose an optimized algorithm for sorting data in real-time systems"). Verbs such as "create" and "design" were used.
- Evaluation: These questions involved making judgments or assessing the effectiveness of different approaches (e.g., "Critique the efficiency of different sorting algorithms"). Verbs like "assess," "evaluate," and "judge" were employed.

Sampling and Validation: The dataset consists of 3691 observations, each representing a generated question (*Text*) and its corresponding Bloom's cognitive level (*Label*). To ensure the validity and alignment of the generated questions, a two-stage validation process was implemented:

- Expert Review: A random sample of questions from each cognitive level was evaluated by subject matter experts specializing in educational assessment and Bloom's Taxonomy. These experts reviewed whether the questions accurately represented the cognitive skill they were intended to measure, offering qualitative feedback on both the content and cognitive alignment.
- Survey Instrument Design: Specific constructs within the survey were developed based on expert recommendations and prior research. Each construct aimed to assess the respondent's perception of question difficulty, relevance, and cognitive demand. The feedback provided by experts allowed for iterative refinement of the prompts, enhancing the consistency and precision of the generated questions across all cognitive levels.

Ethical considerations were important in this study, particularly regarding the potential biases in AI-generated content. We conducted a thorough analysis of the generated questions to ensure they did not perpetuate stereotypes or biases. This involved examining the language and context of the questions for neutrality and inclusivity, and making adjustments to the prompt design where necessary to mitigate any identified biases.

The data is distributed as shown in Table 1.

Table 1. Data distribution

| Bloom's Level | Count |
|---|---|
| Knowledge | 757 |
| Comprehension | 787 |
| Application | 792 |
| Analysis | 455 |
| Synthesis | 449 |
| Evaluation | 451 |
| **Total** | **3691** |

#### B. Calculated Metrics

To ensure that the questions generated by our system align with educational standards and are suitable for their intended Bloom's Taxonomy levels, we calculated several key metrics: Question length (*L*), Flesch-Kincaid Grade Level (*FKGL*), Vocabulary Richness (Type-Token Ratio—*TTR*), and Lexical Density (*LD*). These metrics provide quantitative measures of readability and complexity, which are crucial for evaluating the appropriateness of educational content.

*1)* Question Length (L) is defined as the total number of words in a question. It serves as a preliminary indicator of its complexity. Longer questions may indicate a higher level of complexity or verbosity, while shorter questions tend to be more straightforward. It is expressed as:

$$L = Number\ of\ words\ in\ the\ question \quad (1)$$

---
[1] https://iowaascd.org/all-about-learning/approaches-to-learning/revised-bloom-taxonomy

Analyzing question length helps determine whether the questions adhere to the desired level of conciseness or detail appropriate for the Bloom's Taxonomy level they target.

*2)* Flesch-Kincaid Grade Level is a readability test designed to estimate the U.S. school grade level required to comprehend a text. This metric is calculated using the following formula:

$$FKGL = 0.39\left(\frac{N_w}{N_s}\right) + 11.8\left(\frac{N_s y_l}{N_w}\right) - 15.59 \quad (2)$$

where:
- $N_w$ is the total number of words in the text;
- $N_s$ is the total number of sentences;
- $N_s y_l$ is the total number of syllables.

A lower *FKGL* indicates that the text is easier to read, while a higher level suggests greater complexity. This measure is particularly useful for determining if questions are appropriately challenging for the Bloom's Taxonomy level they are intended to assess. For example, questions targeting higher-order skills such as "Analysis" or "Evaluation" might be expected to have a higher *FKGL* compared to those aimed at "Knowledge."

*3)* Vocabulary Richness (Type-Token Ratio—TTR) is an indicator of the lexical diversity within a text. This metric is calculated using the following formula:

$$TTR = \frac{N_{unique}}{N_w} \quad (3)$$

where:
- $N_{unique}$ is the number of unique words in the text;
- $N_w$ is the total number of words in the text.

A higher TTR indicates greater lexical diversity, meaning a larger proportion of the words in the text are unique. This measure is useful for evaluating the complexity and variety of the vocabulary used in the questions. For instance, questions intended to assess higher-order cognitive skills, such as "Synthesis" or "Evaluation," might exhibit higher TTR values due to the necessity of using more varied and sophisticated language.

*4)* Lexical Density (LD) is a measure of the proportion of content words (nouns, verbs, adjectives, and adverbs) to the total number of words in a text. It provides an indication of the informational content and complexity of the text. This metric is calculated using the following formula:

$$LD = \frac{N_{content}}{N_w} \quad (4)$$

where:
- $N_{content}$ is the number of content words in the text;
- $N_w$ is the total number of words in the text.

Content words are those that carry significant meaning and are essential for understanding the text, excluding common stop words such as "the," "is," and "and." A higher *LD* indicates a text rich in meaningful words, which is often associated with higher complexity. This measure is particularly valuable for determining the appropriateness of questions for different Bloom's Taxonomy levels. Questions designed to assess complex cognitive processes, such as "Analysis" or "Evaluation," are expected to have higher *LD* compared to questions aimed at lower-order skills like "Knowledge."

*C. Classification Using Multinomial Logistic Regression*

Multinomial logistic regression is a statistical technique used for classifying observations into multiple categories by estimating the probabilities of each class based on the input features [34]. This is an extension of binary logistic regression to handle cases where there are more than two possible outcomes, providing a way to model the relationship between categorical response variables and predictor variables. In this paper, we used multinomial regression to predict the question label based on several independent variables mentioned in the previous subsection.

*D. Classification Using Naive Bayes*

Naive Bayes is a probabilistic classification technique based on Bayes' Theorem, which provides a mathematical framework for updating the probability of a hypothesis as more evidence becomes available. The "naive" aspect of the algorithm assumes that all input features are independent of each other, which simplifies the computation but may not fully capture the complexity of real-world data [35]. Despite this simplifying assumption, Naive Bayes is widely used for text classification due to its efficiency and effectiveness, especially in high-dimensional datasets with sparse features. In this study, Naive Bayes was applied to classify questions into categories based on their textual content, leveraging its strength in handling multiclass problems and providing a straightforward probabilistic interpretation of the results.

*E. Classification Using Linear Support Vector Classification (SVC)*

Support Vector Machines (SVM) are a powerful set of supervised learning methods used for classification and regression tasks. Linear Support Vector Classification (SVC) is a variant of SVM that seeks to find a hyperplane in a high-dimensional space that best separates the classes, maximizing the margin between data points of different categories [36]. The linear nature of SVC makes it particularly suitable for cases where the classes are linearly separable or nearly so. This method is known for its robustness in handling high-dimensional data and providing clear decision boundaries. In the context of this study, Linear SVC was employed to classify questions by learning the optimal hyperplane that distinguishes between different cognitive categories, relying on the textual features of the questions.

*F. Classification Using Transformer-Based Deep Learning Model*

Transformers represent a significant advancement in natural language processing, using self-attention mechanisms to capture long-range dependencies and contextual information within text sequences [37]. DistilBERT, a variant of the original BERT model, is a distilled, smaller version designed to retain much of the performance while being more efficient in terms of speed and computational resources. Transformers have been particularly successful in tasks requiring deep understanding of language semantics and context, making them highly effective for text classification problems. In this research, DistilBERT was fine-tuned to classify questions into different cognitive categories as defined by Bloom's Taxonomy.

### G. Hypotheses

To address the research questions, we propose the following hypotheses:

H1: AI-generated questions will accurately reflect lower-order cognitive levels (Knowledge, Comprehension, and Application) when categorized by traditional machine learning models such as Multinomial Logistic Regression, Naive Bayes and Support Vector Machines.

Prior studies have demonstrated the efficacy of Naive Bayes in classifying questions into Bloom's Taxonomy levels. For instance, Naive Bayes with Laplace Smoothing improved accuracy by 39.6% when applied to high school biology questions, achieving up to 75.94% accuracy when tested with mixed data samples [38]. Similarly, Naive Bayes achieved a classification accuracy of 91% when categorizing quiz questions according to Bloom's revised cognitive levels [39]. These findings support the hypothesis that traditional classifiers like Naive Bayes can perform well in categorizing questions into lower-order cognitive tasks.

H2: Traditional classifiers like Naive Bayes will struggle with higher-order cognitive levels (Analysis, Synthesis, Evaluation), while advanced models such as Support Vector Machines and deep learning methods will improve classification accuracy for these levels.

Studies comparing Naive Bayes with other classifiers, such as Support Vector Machines (SVM), have shown that SVM performs better in classifying questions into higher-order cognitive levels. In one study, SVM with SMOTE achieved a 98% accuracy, significantly outperforming Naive Bayes, which reached 91% [39]. This supports our hypothesis that Naive Bayes, while useful for lower-order tasks, may not be sufficient for complex cognitive levels. Another study demonstrated that normalizing TF-IDF variants further enhances SVM's accuracy for higher-level questions, which supports our methodology of including advanced classification models [40]. Additionally, research on classification using modified TF-IDF and word2vec methods achieved significant results, showing improvements in classification performance with features like TFPOS-IDF and pre-trained word2vec [41]. Additionally, research on automatic classroom question classification using TF-IDF features reported an accuracy of 86% [42]. This further supports the inclusion of advanced classification models.

H3: Deep learning models, such as LSTM and CNN, will significantly enhance classification accuracy, especially for complex cognitive tasks involving higher-order levels of Bloom's Taxonomy.

Recent studies have shown the advantages of deep learning techniques in classifying questions into Bloom's Taxonomy [43, 44]. For example, Long Short-Term Memory (LSTM) models achieved 87% accuracy in classifying learning outcomes into Bloom's cognitive levels, with particular improvement in higher-order classifications [45]. Convolutional Neural Networks (CNN) have similarly demonstrated success in categorizing questions across various cognitive levels, achieving high levels of precision and recall [46]. This evidence supports our hypothesis that deep learning models are more capable of handling the complexity of higher-order cognitive levels, providing significant improvements over traditional methods.

## IV. EXPERIMENTS & RESULTS

This section presents the results of experiments designed to evaluate the effectiveness of AI-generated questions. To test our hypotheses, we employed Multinomial Logistic Regression, Naive Bayes, Support Vector Machines (SVM), and transformer-based models (DistilBERT) to classify these questions according to Bloom's Taxonomy. These models were chosen for their proven efficacy in text classification tasks, particularly within educational contexts. We conducted a detailed analysis and comparison of these models, measuring performance using accuracy, F1-score, and other classification metrics. This analysis focuses on each model's ability to categorize questions across both lower- and higher-order Bloom's levels, providing insights into the alignment of AI-generated questions with Bloom's Taxonomy. The results directly address the three hypotheses proposed at the outset of this study.

### A. Exploratory Results

The analysis of our experiments revealed several important patterns. As observed in Fig. 1., there is a clear trend where the length of questions increases with the Bloom's level. This pattern reflects the nature of higher cognitive processes, which often require more elaborate and detailed questions. For example, questions designed to evaluate higher-order skills such as "Analysis" or "Evaluation" generally involve multiple facets of information, thereby increasing their length. This trend aligns with our expectations, as higher Bloom's levels involve more complex cognitive demands.

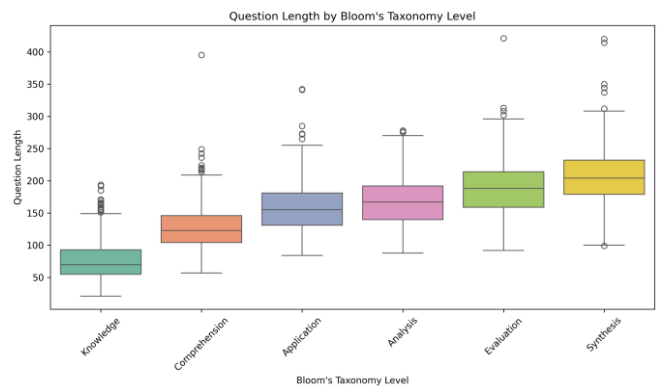

Fig. 1. QL by Bloom's taxonomy level.

In addition to question length, FKGL and LD also show a general increase with higher Bloom's levels, although this trend is less pronounced for LD as shown in Figs. 2 and 3. respectively. Higher FKGL values suggest that questions targeting advanced cognitive levels are more challenging in terms of readability, consistent with the complexity expected at these levels. Similarly, higher LD indicates a greater proportion of content words, reflecting increased informational complexity in these questions.

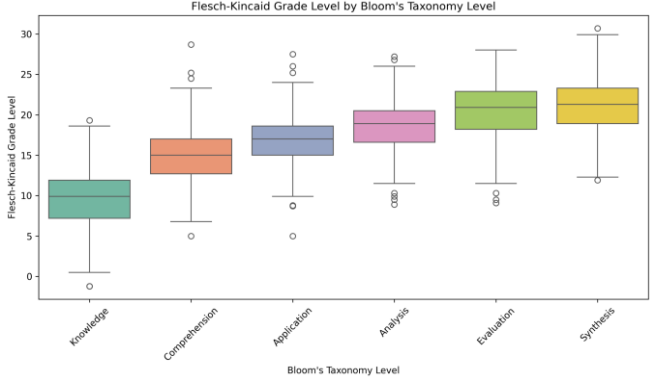

Fig. 2. FKGL by Bloom's taxonomy level.

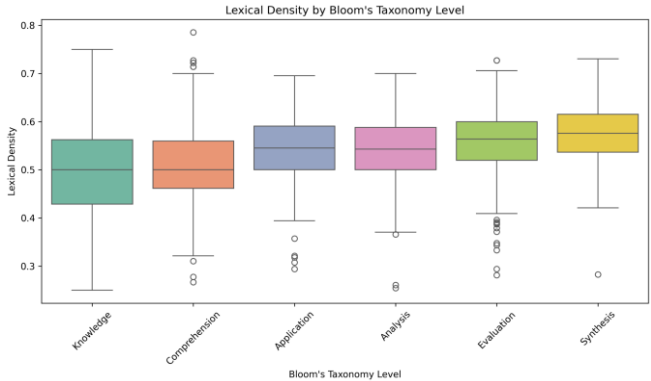

Fig. 3. LD by Bloom's taxonomy level.

Interestingly, as shown in Fig .4., no significant correlation was found between Type-Token Ratio (TTR) and Bloom's levels. This suggests that while questions at higher cognitive levels may involve more detailed content, the diversity of vocabulary used does not necessarily follow a predictable pattern.

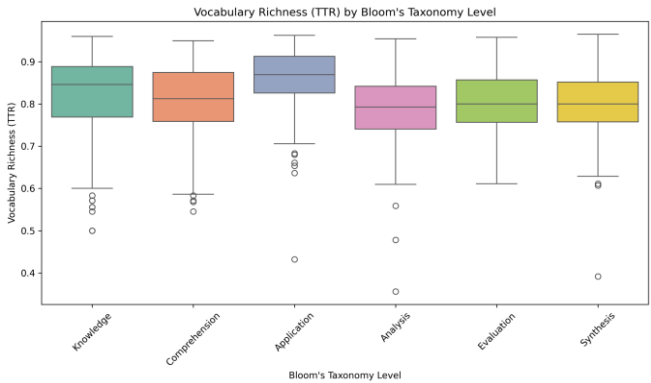

Fig. 4. TTR by Bloom's taxonomy level.

Furthermore, a strong positive correlation of 0.84 between question length and FKGL was identified, indicating that longer questions are generally associated with a higher readability grade level. This supports the notion that more detailed questions tend to be more complex.

### B. Multinomial Logistic Regression Classification Performance

In Experiment 1, we employed multinomial logistic regression to classify questions into the six Bloom's Taxonomy levels using a balanced dataset. The multinomial logistic regression model aimed to predict the cognitive level of each question based on its features. The results, shown in Table 2, revealed significant variability in the model's performance across different cognitive levels. Specifically, the model achieved the highest accuracy for the "Knowledge" level, with an F1-score of 0.83. This indicates that the model was particularly effective at classifying straightforward recall tasks. Conversely, the performance for higher-order cognitive skills such as "Synthesis" and "Analysis" was notably lower, with F1-scores of 0.44 and 0.52, respectively. This variability highlights the model's challenges in accurately classifying questions that require more complex cognitive processing.

Table 2. Performance rates for each class for Experiment 1

| Level | Precision | Recall | F1-score |
|---|---|---|---|
| Analysis | 0.54 | 0.50 | 0.52 |
| Application | 0.46 | 0.53 | 0.49 |
| Comprehension | 0.64 | 0.67 | 0.65 |
| Evaluation | 0.60 | 0.60 | 0.60 |
| Knowledge | 0.84 | 0.82 | 0.83 |
| Synthesis | 0.46 | 0.42 | 0.44 |
| Accuracy | | | 0.59 |

These findings partially support Hypothesis 1, as traditional models like multinomial logistic regression effectively classify lower-order tasks but are less effective for higher-order levels, confirming Hypothesis 2.

### C. Merging Higher-Level Bloom's Categories

Building on the insights from Experiment 1, Experiment 2 involved a strategic adjustment where we merged the three highest levels of Bloom's Taxonomy—Analysis, Evaluation, and Synthesis—into a single class labeled "Higher-Order" [2, 8]. This approach was designed to simplify the classification task by reducing the number of categories and potentially enhancing the model's accuracy. We retrained the multinomial logistic regression model with this revised category structure.

The results, shown in Table 3., from this experiment showed a marked improvement in classification accuracy. The F1-score for the merged "Higher-Order" class increased to 0.68, reflecting the model's enhanced ability to classify complex cognitive tasks more effectively after the consolidation. The "Knowledge" class continued to perform well, with an F1-score of 0.83, demonstrating that the model's proficiency in handling lower-order questions remained strong. Improvements were also observed in the F1-scores for the "Application" and "Comprehension" categories, suggesting that the simplified classification task positively impacted the accuracy for these intermediate cognitive skills. Overall, the model's accuracy increased to 0.68, indicating that merging higher-level categories effectively improved performance by reducing classification complexity.

Table 3. Performance rates for each class for Experiment 2

| Level | Precision | Recall | F1-score |
|---|---|---|---|
| Higher-Order | 0.72 | 0.65 | 0.68 |
| Application | 0.59 | 0.64 | 0.61 |
| Comprehension | 0.59 | 0.63 | 0.61 |
| Knowledge | 0.85 | 0.81 | 0.83 |

| Level | Precision | Recall | F1-score |
|---|---|---|---|
| Accuracy | | | 0.68 |

### D. Further Refinement and Performance

In Experiment 3, we further refined the approach by consolidating "Comprehension" and "Application" into a single "Mid-Order" class. This refinement aimed to optimize the classification of these cognitive levels even further using multinomial logistic regression.

The results, as shown in Table 4, demonstrated significant improvements in performance metrics. The "Higher-Order" class, now representing the combined higher-order levels, achieved an F1-score of 0.78. This indicates a substantial enhancement in the model's ability to handle complex cognitive tasks after refining the merging strategy. The "Knowledge" class also saw a slight increase in its F1-score to 0.86, reflecting the model's continued strength in identifying recall-based questions. The "Mid-Order" category improved to an F1-score of 0.65, suggesting that the refined merging approach positively impacted the classification of application-based questions. Overall, the model's accuracy rose to 0.76, demonstrating that the refined consolidation strategy not only improved performance for higher-level categories but also enhanced the model's overall classification capabilities.

Table 4. Performance rates for each class for Experiment 3

| Level | Precision | Recall | F1-score |
|---|---|---|---|
| Higher-Order | 0.81 | 0.76 | 0.78 |
| Mid-Order | 0.62 | 0.65 | 0.65 |
| Knowledge | 0.86 | 0.86 | 0.86 |
| Accuracy | | | 0.75 |

### E. Naive-Bayes Algorithm for Multi-Class Classification

In Experiment 4, we applied the Naive-Bayes algorithm, a common choice for multi-class classification problems, to our dataset. The Naive-Bayes classifier achieved an overall accuracy of 0.79 as shown in Table 5. It demonstrated high precision and recall for the "Knowledge" class, with an F1-score of 0.95, indicating strong performance in classifying basic recall questions. The "Application" class also showed good results with an F1-score of 0.89. However, the model struggled with the higher-order cognitive levels, such as "Analysis" and "Synthesis," which had lower F1-scores of 0.66 and 0.62, respectively. This suggests that while the Naive-Bayes algorithm performs well with simpler categories, its effectiveness diminishes for more complex cognitive tasks.

Table 5. Performance rates for each class for Experiment 4

| Level | Precision | Recall | F1-score |
|---|---|---|---|
| Knowledge | 0.97 | 0.93 | 0.95 |
| Comprehension | 0.77 | 0.80 | 0.78 |
| Application | 0.88 | 0.90 | 0.89 |
| Analysis | 0.68 | 0.64 | 0.66 |
| Synthesis | 0.67 | 0.58 | 0.62 |
| Evaluation | 0.62 | 0.74 | 0.68 |
| Accuracy | | | 0.79 |

These results reinforce Hypothesis 1 by demonstrating Naive Bayes' efficacy for lower-order tasks, but also support Hypothesis 2, indicating that traditional models falter at higher cognitive levels.

### F. Linear Support Vector Classifier (SVC) for Multi-Class Classification

Experiment 5 employed a Linear Support Vector Classifier (SVC) to tackle the multi-class classification problem. The SVC achieved a notable accuracy of 0.83. The model excelled in classifying "Knowledge" and "Application" levels, with F1-scores of 0.92 and 0.93 respectively as shown in Table 6. It also showed improved performance for "Comprehension" and "Evaluation," with F1-scores of 0.80 and 0.76. However, the "Analysis" and "Synthesis" levels still presented challenges, with F1-scores of 0.70 and 0.66. This suggests that while SVC is effective for many cognitive levels, further refinements may be needed to enhance its performance for higher-order tasks.

Table 6. Performance rates for each class for Experiment 5

| Level | Precision | Recall | F1-score |
|---|---|---|---|
| Knowledge | 0.91 | 0.94 | 0.92 |
| Comprehension | 0.75 | 0.85 | 0.80 |
| Application | 0.93 | 0.94 | 0.93 |
| Analysis | 0.78 | 0.64 | 0.70 |
| Synthesis | 0.77 | 0.58 | 0.66 |
| Evaluation | 0.73 | 0.78 | 0.76 |
| Accuracy | | | 0.83 |

These results continue to validate Hypothesis 2: although SVC improves performance over Naive Bayes, it still struggles to classify higher-order tasks effectively, particularly those requiring synthesis and evaluation.

### G. Deep Learning with DistilBERT

In Experiment 6, we utilized DistilBERT, a transformer-based model, to handle the classification task. DistilBERT, known for its efficiency and speed compared to the original BERT model, was fine-tuned on our dataset. The deep learning model achieved a validation accuracy of 0.91 and a validation loss of 0.28. The confusion matrix shows that the model performed exceptionally well across all levels, with high precision for "Knowledge" (0.97) and "Comprehension" (0.88). The model also demonstrated improvements in the higher-order levels, with F1-scores of 0.79 for "Analysis" and 0.84 for "Synthesis". The results from this experiment, shown in Table 7, highlight the superior performance of deep learning models in understanding and classifying complex cognitive tasks compared to traditional methods.

Table 7. Performance rates for each class for Experiment 6

| Level | Precision | Recall | F1-score |
|---|---|---|---|
| Knowledge | 0.97 | 0.96 | 0.97 |
| Comprehension | 0.88 | 0.90 | 0.89 |
| Application | 0.94 | 0.91 | 0.93 |
| Analysis | 0.80 | 0.78 | 0.79 |
| Synthesis | 0.86 | 0.82 | 0.84 |
| Evaluation | 0.87 | 0.86 | 0.87 |
| Accuracy | | | 0.91 |

These findings fully support Hypothesis 3, confirming that deep learning models outperform traditional classifiers across all levels of Bloom's Taxonomy, particularly in handling higher-order cognitive processes.

## V. DISCUSSION

The findings of this study provide significant insights into the performance of AI-driven models in generating and classifying Multiple-Choice Questions (MCQs) according to Bloom's Taxonomy. Our results are consistent with earlier studies but also offer new perspectives on the challenges and opportunities in aligning AI-generated content with established educational frameworks. By comparing the performance of traditional machine learning models (Multinomial Logistic Regression, Naive Bayes, and Linear Support Vector Classification) and a transformer-based deep learning model (DistilBERT), we observed trends that both corroborate and challenge prior research.

Table 8. Comparative results

| Experiment | Method | Accuracy | Notes |
|---|---|---|---|
| Exp 1 | Multinomial Logistic Regression | 0.63 | Baseline model, struggled with higher-order levels like Synthesis and Analysis. |
| Exp 2 | Multinomial Logistic Regression (Merged Categories) | 0.68 | Merged Analysis, Synthesis, and Evaluation; improved performance. |
| Exp 3 | Multinomial Logistic Regression (Refined Merged Categories) | 0.75 | Further merging; better accuracy and F1-scores, especially for Knowledge and Mid-Order. |
| Exp 4 | Naive-Bayes | 0.79 | High performance in lower-order levels, struggled with higher-order tasks. |
| Exp 5 | Linear Support Vector Classifier | 0.83 | Strong performance overall, challenges with Analysis and Synthesis levels. |
| Exp 6 | DistilBERT (Deep Learning) | 0.91 | Highest accuracy, excellent across all levels, particularly effective with higher-order tasks. |

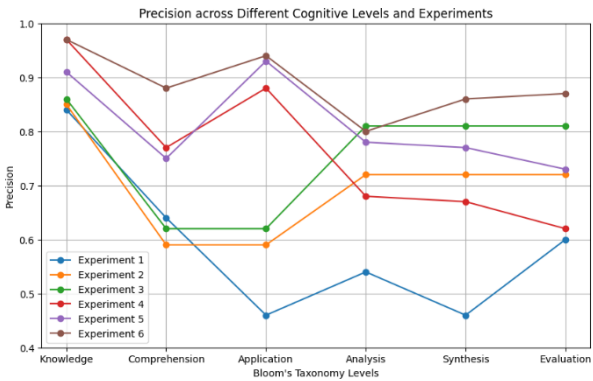

Fig. 5. Precision across different cognitive levels and experiments.

The results, as shown in Table 8 and Fig. 5, indicate that while the AI model could generate questions that align well with lower-order cognitive levels, such as Knowledge and Comprehension, it struggled with higher-order levels, including Analysis, Synthesis, and Evaluation. The precision of classification varied significantly across different levels, with lower-order levels achieving higher accuracy. This discrepancy suggests that while AI can generate content that matches the surface features of Bloom's Taxonomy, capturing the deeper, more abstract cognitive skills required for higher-order levels remains a challenge. This points to a need for further refinement in AI models to improve their capability to generate and classify questions that truly reflect the complexity of higher-order thinking.

A. Main Findings and Comparative Analysis

The most notable finding of this study is the superior performance of the transformer-based model, DistilBERT, across all Bloom's Taxonomy levels, particularly in higher-order cognitive tasks (Analysis, Synthesis, and Evaluation). This result aligns with prior studies that have demonstrated the advantages of deep learning models in capturing contextual and semantic nuances in text-based tasks [45]. For instance, recent work on LSTM and CNN models supports our findings, showing that deep learning techniques significantly outperform traditional models in complex cognitive classifications [43]. However, our study extends these insights by applying them specifically to the context of AI-generated MCQs, reinforcing the idea that deep learning can enhance the educational content generation process.

In contrast, traditional models performed well in classifying lower-order cognitive tasks, such as Knowledge and Comprehension. This is consistent with previous research, which has shown that simpler machine learning algorithms are sufficient for classifying recall-based tasks. For example, the Naive Bayes classifier achieved high accuracy for the "Knowledge" level, which mirrors findings from earlier studies in text classification [38, 39]. However, these models struggled with higher-order levels, as expected, due to their reliance on surface-level lexical patterns rather than deep semantic understanding.

Table 8 compares the results across these experiments and reveals several insights. Multinomial logistic regression, while effective, showed variability in performance, particularly with higher cognitive levels. The consolidation strategies improved performance but still left room for further refinement. In contrast, the Naive-Bayes algorithm provided high accuracy for lower-order tasks but struggled with higher-order levels. The Linear SVC demonstrated strong overall performance but faced challenges with complex cognitive classifications. Finally, the deep learning approach with DistilBERT significantly outperformed traditional methods, achieving the highest accuracy and demonstrating strong performance across all Bloom's levels. This comparison further supports Hypothesis 3 and highlights the need for advanced deep learning techniques to classify complex cognitive levels effectively.

The performance gap between traditional and deep learning models suggests that while traditional models can serve as a baseline for simple cognitive tasks, advanced models are essential for more complex assessments. This finding is particularly important in educational contexts where promoting higher-order thinking is a key objective. AI tools must be capable of generating questions that reflect not only surface-level cognitive processes but also deeper, more abstract cognitive skills.

B. Relation to Previous Studies

The challenges encountered by traditional models in classifying higher-order tasks align with prior research, particularly studies that emphasize the limitations of machine learning models in handling complex educational content. For instance, studies using Naive Bayes and SVMs have reported similar difficulties in distinguishing between higher-order cognitive processes, such as Analysis and Synthesis [39]. These models often rely on word frequency and simple feature extraction methods, which are not adequate for capturing the intricacies of questions designed to assess critical thinking and problem-solving.

However, our study contributes to the literature by demonstrating that combining or merging higher-order categories (as we did in experiments 2 and 3) can moderately improve classification accuracy. While this approach does not fully address the complexities of higher-order cognitive tasks, it suggests that reducing the classification burden for traditional models can yield marginal performance gains. This finding offers a practical recommendation for improving traditional classification methods, although it underscores the need for more sophisticated models.

Furthermore, our results support the growing body of literature advocating for the integration of AI with structured educational frameworks such as Bloom's Taxonomy. Recent studies, including those on GPT-4's potential in generating contextually relevant questions, have highlighted the importance of aligning AI-generated content with established cognitive frameworks [31]. Our study adds to this conversation by empirically validating the effectiveness of such integrations, particularly in enhancing the pedagogical value of AI-generated assessments.

### C. Critical Reflection and Limitations

While our findings are encouraging, it is essential to recognize the limitations of the current study. One key limitation is the reliance on pre-defined prompts and action verbs for question generation. As noted in previous studies, the choice of action verbs can significantly influence the quality and cognitive level of the generated questions. Although we selected verbs based on established references, it is possible that the variability in verb selection may have introduced biases in the classification outcomes. Future research could explore the use of more dynamic prompt engineering techniques to mitigate this issue.

Another important consideration is the representativeness of the dataset. Our dataset focused on questions within the domain of computer science, which may limit the generalizability of the findings to other disciplines. Previous research has shown that domain-specific features can impact the performance of AI models in educational contexts [4]. Therefore, it would be valuable to extend this research to other fields of study to determine whether the observed trends hold across different subject areas.

Finally, although DistilBERT outperformed traditional models, it is important to note that even advanced models exhibited challenges in fully capturing the complexity of higher-order cognitive tasks. As shown in our analysis, the F1-scores for higher-order levels, while improved, were still lower than those for lower-order levels. This suggests that further refinement of AI models is necessary to enhance their ability to generate and classify questions that target deep, abstract thinking. Future research could investigate hybrid models that combine the strengths of both traditional and deep learning approaches.

### D. Implications for Educational Practice

Thus, the integration of Bloom's Taxonomy into AI-driven tools like OneClickQuiz not only enhances the alignment of questions with educational objectives but also offers educators a powerful resource for designing assessments that promote higher-order thinking. This capability allows for more targeted assessments, ensuring that students are evaluated across a range of cognitive skills, from basic recall to complex analysis and synthesis. Furthermore, the ability to automatically generate such questions can significantly reduce the time and effort required for educators to develop comprehensive assessments, allowing them to focus more on personalized instruction and feedback.

The implications of this study are significant for educators and developers of AI-driven educational tools. By integrating Bloom's Taxonomy into AI-based question generation, educational technologies can better align with cognitive objectives that promote higher-order thinking. However, our findings also suggest that educators should be cautious when relying solely on traditional machine learning models for content generation, particularly for complex tasks. Instead, advanced models such as transformers should be prioritized for tasks that require deep cognitive engagement.

For AI to truly transform educational practices, further advancements in model architecture and prompt engineering will be required. By addressing these challenges, AI-driven tools like OneClickQuiz can play a pivotal role in promoting critical thinking and deeper learning in educational environments.

### E. Extending the Generalizability of the Proposed Framework

The methodologies and classification models developed in this study, while initially applied to OneClickQuiz within the Moodle platform, are inherently adaptable to a wide range of educational contexts and tools. The underlying framework leverages Bloom's Taxonomy as a universal structure for cognitive classification, enabling its deployment across various domains and platforms. The machine learning models, particularly the Transformer-based architecture of DistilBERT, can be fine-tuned for diverse datasets and subject areas, allowing for scalable implementation. This generalizability positions the framework as a flexible solution for aligning AI-generated questions with cognitive objectives, enhancing its potential impact on automated assessments and personalized learning in broader educational settings.

## VI. Conclusion

In this study, we investigated the alignment of AI-generated questions with Bloom's Taxonomy, using data created through Google's Vertex AI Text Generation Model. Our findings revealed that while the AI model was effective in generating questions that adhered to the surface characteristics of lower-order cognitive levels, it faced challenges in accurately aligning questions with higher-order cognitive skills. Through a series of classification experiments, we demonstrated that the precision of AI models varied across cognitive levels, with advanced models like DistilBERT showing notable improvements, yet still encountering difficulties in fully capturing the complexities of higher-order thinking.

The main contribution of this research lies in the integration of Bloom's Taxonomy into AI-driven assessment tools, providing insights into how AI models can be adapted to educational frameworks. This work underscores the potential of AI in automating quiz creation, particularly by tailoring content to specific cognitive objectives. However, a key limitation of this study is the model's limited ability to consistently align with higher-order cognitive levels, such as Synthesis and Evaluation, which highlights the need for further refinement in AI training and modeling techniques.

Future research should address these limitations by focusing on enhancing AI algorithms and expanding training datasets to better support the accurate generation and classification of questions across all cognitive levels. Moreover, the development of hybrid AI models or more advanced language understanding mechanisms could improve the pedagogical effectiveness of AI-generated educational content, making it more adaptable to the nuances of higher-order cognitive tasks. By addressing these gaps, AI-driven educational technologies can further contribute to the advancement of automated assessment tools that foster critical thinking and deeper learning.

To extend the applicability of our findings, we also propose the development of a generic AI-driven question classification framework. This framework would integrate the principles outlined in this study and be adaptable to various learning management systems (LMS) and standalone applications. By designing a modular architecture, the framework can accommodate different cognitive frameworks, including Bloom's Taxonomy, Webb's Depth of Knowledge, and others. The core components would include a model selection module, customizable prompts, and a feedback mechanism for continuous improvement. Such a system would support educators in creating and categorizing questions dynamically, ensuring pedagogical relevance across disciplines and learning contexts.